%% file: main.tex
\documentclass[10pt,twocolumn,letterpaper]{article}

\usepackage{cvpr}              


\definecolor{cvprblue}{rgb}{0.21,0.49,0.74}
\usepackage[pagebackref,breaklinks,colorlinks,allcolors=cvprblue]{hyperref}

\usepackage{amsmath}
\usepackage{graphicx}
\usepackage{enumitem}
\usepackage{makecell}
\usepackage{multirow}
\usepackage{arydshln}
\usepackage{subcaption}
\usepackage{amssymb}
\usepackage{pifont}
\newcommand{\cmark}{\ding{51}}%
\newcommand{\xmark}{\ding{55}}%

\usepackage{array}
\newcolumntype{$}{>{\global\let\currentrowstyle\relax}}
\newcolumntype{^}{>{\currentrowstyle}}
\newcommand{\rowstyle}[1]{\gdef\currentrowstyle{#1}%
  #1\ignorespaces
}

\def\paperID{4193} 
\def\confName{CVPR}
\def\confYear{2025}

\title{PointRecon: Online Point-based 3D Reconstruction via Ray-based 2D-3D Matching}

\author{Chen Ziwen$^1$~~~
Zexiang Xu$^2$~~~
Li Fuxin$^1$~~~
\\\\
$^1$Oregon State University~~~
$^2$Adobe Research\\
}

\begin{document}
\maketitle
\input{sec/0_abstract}

\input{sec/1_intro}
\input{sec/2_related}

\input{sec/3_method}

\input{sec/4_experiments}

\input{sec/5_conclusion}

{
    \small
    \bibliographystyle{ieeenat_fullname}
    \bibliography{main}
}


\end{document}

%% file: sec/0_abstract.tex
\begin{abstract}
We propose a novel online, point-based 3D reconstruction method from posed monocular RGB videos. Our model maintains a global point cloud representation of the scene, continuously updating the features and 3D locations of points as new images are observed. It expands the point cloud with newly detected points while carefully removing redundancies. The point cloud updates and the depth predictions for new points are achieved through a novel ray-based 2D-3D feature matching technique, which is robust against errors in previous point position predictions. In contrast to offline methods, our approach processes infinite-length sequences and provides real-time updates. Additionally, the point cloud imposes no pre-defined resolution or scene size constraints, and its unified global representation ensures view consistency across perspectives. Experiments on the ScanNet dataset show that our method achieves comparable quality among online MVS approaches.
Project page: \url{https://arthurhero.github.io/projects/pointrecon/}
\end{abstract}

%% file: sec/1_intro.tex
\section{Introduction}
\label{sec:intro}

\begin{figure*}[h]
    \centering
    \includegraphics[width=0.8\linewidth]{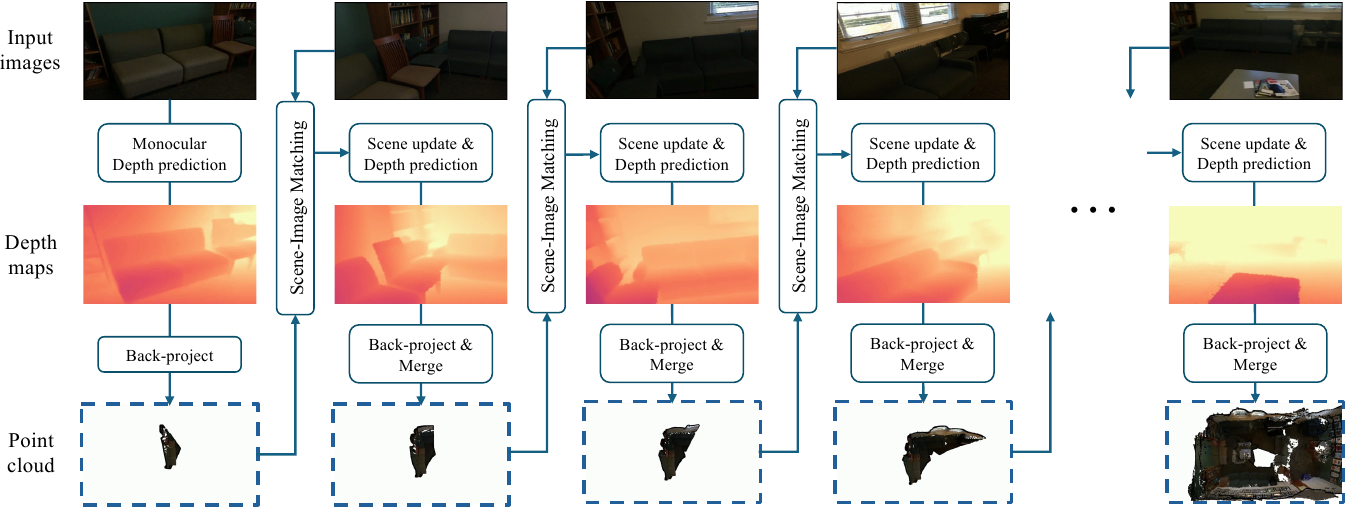}
    \caption{\small \textbf{Workflow of PointRecon.} We begin with monocular depth prediction for the first image, lifting 2D points into 3D space to form the initial point cloud. For each subsequent image, we perform feature matching between the 2D image features and the 3D point cloud features to update the features and positions of the point cloud and to predict depth for the 2D image. Finally, the new points are merged with the existing point cloud.}
    \vskip -0.1in
\end{figure*}

3D reconstruction is a fundamental problem in computer vision. The ability to recover the 3D geometry of a scene from a set of posed RGB images, known as multi-view stereo (MVS), enables numerous applications, including object retrieval, robotics, and computer-aided 3D design. Unlike methods relying on expensive depth sensors such as LiDAR, RGB-based MVS offers a cost-effective alternative, eliminating calibration errors and synchronization issues associated with multiple sensors. Among other reconstruction scenarios, online MVS methods are particularly well-suited for reconstructing large scenes from extremely long image sequences.


Deep learning has enabled a wide range of MVS methods. Depth prediction-based approaches~\citep{gcnet,psmnet,mvsnet,mvdepthnet,deepmvs,deepvideomvs,dpsnet,simplerecon} accumulate features from multiple views into a per-view cost volume to regress depth maps, while volumetric methods~\citep{mvsmachine,atlas,neuralrecon,vortx,transformerfusion,visfusion} aggregate features into a global voxel grid to predict occupancy or TSDF values. Point cloud-based methods~\citep{quasi,pointmvs} directly estimate surface point positions. However, depth prediction methods regress depth maps independently for each image, which can lead to inconsistencies across views. Volumetric methods maintain a global voxel grid, but require predefined voxel sizes and suffer from high memory costs at high resolutions, resulting in coarse meshes. Additionally, they are limited to bounded areas and cannot extend to an infinite horizon, despite the advantages of using cameras.

A globally consistent 3D point cloud representation addresses some of these drawbacks. It is a sparse representation that significantly reduces memory consumption compared to volumetric methods. Furthermore, point clouds do not require predefined voxel sizes and can adaptively represent surface details with varying densities, allowing for greater detail in areas that need it. However, existing point cloud-based MVS methods typically rely on iterative optimization~\citep{pointmvs,gaussian}, making them unsuitable for online algorithms.

In this paper, we propose PointRecon, an online point-based MVS method that maintains a global feature-augmented point cloud for scene representation. When a new image is observed, we match its 2D features with existing 3D points using our novel ray-based 2D-3D matching technique, which is robust against errors in previous position predictions. 
We begin with matching each 3D feature in the point cloud with the image features by projecting the 3D camera ray onto the image plane, similar to an epipolar line. The 3D feature is then compared against the 2D features along the projected ray, followed by an update of its feature and position. Next, we match each 2D feature with the existing point cloud by extending its corresponding camera ray into the 3D space. The 2D feature is similarly compared against the 3D features along this ray, informing the next depth prediction step. Note that since point clouds are compact representations, they contain features only on the predicted surface, which may sometimes be in incorrect location. Naively searching for the nearest 3D points along the camera ray risks missing true matches. However, the camera rays of a matching pair must intersect. Thus, we enhance robustness by gathering 3D neighbors based on the minimal ray distance to the target 2D feature's camera ray.
Finally, we merge the new points from the image into the existing point cloud using a carefully designed merging module that effectively removes redundant points. We also provide simple algorithms to render depth maps and fuse 3D meshes from the point cloud.

It is important to note that the scope of online MVS methods differs from that of SLAM methods~\citep{orb,droid,go,nicer,mod}. Our method recovers geometry from a sequence of posed images and operates purely in a feed-forward manner, while SLAM methods estimate both camera pose and geometry, typically involving optimization steps like the Gauss-Newton algorithm and bundle adjustment.

In summary, our contributions are:
\begin{itemize}
\vspace{-0.03in}
\item We propose an online point-based MVS method that maintains a feature-augmented 3D point cloud, \textbf{unconstrained by input sequence length, predefined voxel resolutions or scene sizes, while ensuring consistency among viewpoints}.
\vspace{-0.02in}
\item We introduce a robust ray-based 2D-3D matching technique for aligning the 3D point cloud with 2D points from incoming images for position estimation. 
\vspace{-0.02in}
\item Our results on the popular ScanNetv2 dataset demonstrate that our approach achieves comparable quality to existing online methods while providing the flexibility mentioned above.
\vspace{-0.05in}
\end{itemize}



%% file: sec/2_related.tex
\section{Related Work}

We summarize previous MVS methods, categorizing them based on their surface representations: depth maps, volumetric TSDF grids, and point clouds.

\textbf{Depth Prediction via Multi-view Stereo Matching.} Depth can be inferred from image patch similarity between two posed cameras viewing the same object~\citep{dtam, remode}, but manually designed measures are often unreliable. CNN-based methods~\citep{cnnstereo, cnnstereo2} improve this by learning patch similarity, though they still lack global context and require post-processing. GC-Net~\citep{gcnet} introduces cost volumes and 3D convolutions for global context, while PSM-Net~\citep{psmnet} further improves it with multi-scale feature maps. DeepMVS~\citep{deepmvs} and MVSNet~\citep{mvsnet}, both extend stereo matching to multiple views, with DeepMVS using plane-sweep volumes and MVSNet introducing a camera frustum-based 3D cost volume. MVDepthNet~\citep{mvdepthnet} further refines this process by compressing features into a single volume for efficient 2D convolution. DPSNet~\citep{dpsnet} improves results by matching deeper features, DeepVideoMVS~\citep{deepvideomvs} integrates reconstruction history with a recurrent network, and SimpleRecon~\citep{simplerecon} further enhances depth map quality by incorporating geometric metadata into the cost volume. While depth map-based MVS methods break down large scenes for faster inference, without a unified scene representation, the independently predicted depth maps fail to ensure consistent views.

\noindent \textbf{Volumetric TSDF Regression.} Instead of predicting depth maps, some methods directly generate global surfaces from cost volumes. LSM~\citep{mvsmachine} back-projects features into a global voxel grid for each image and fuses them to regress global voxel occupancy. Atlas~\citep{atlas} extends this to scenes by accumulating features into a global voxel grid and using 3D convolution to regress TSDF values. NeuralRecon~\citep{neuralrecon} introduces an online approach that incrementally constructs and fuses local grids with a global grid using a GRU, while also sparsifying fine grids based on coarse predictions. Building on this, TransformerFusion~\citep{transformerfusion} leverages transformers to selectively fuse image features into the voxel grid. Further improvements are made by VisFusion~\citep{visfusion}, which incorporates ray-based sparsification, and DG-Recon~\citep{dgrecon}, which enhances the process with monocular depth priors. More recent offline methods like VoRTX~\citep{vortx} incorporate ray direction and depth for view-aware attention, while FineRecon~\citep{finerecon} and CVRecon~\citep{cvrecon} enhance results through fine-grained supervision and cost-volume integration. While volumetric methods offer consistency and can infer unseen surfaces, they rely on predefined grids (often 4cm in size), limiting their ability to capture finer details.


\noindent \textbf{Point Cloud-based Reconstruction.} 
Early methods like~\citep{quasi} extract keypoints, back-project them into 3D space to form a point cloud, and optimize a surface over it. Deep learning-based approaches like Point-MVSNet~\citep{pointmvs} predict a coarse depth map, back-project to a point cloud, augment it with multi-view features, and iteratively refine point positions or depth maps. However, these methods are not online and focus primarily on object datasets. To the best of our knowledge, our method is the first online point-based MVS approach generalized to large scenes. Recently, Gaussian splatting-based reconstruction methods have gained attention, where each Gaussian can be viewed as a point with scale and opacity. 
While the original 3D GS~\citep{gaussian} requires per-scene optimization, recent methods have extended it to feed-forward reconstruction \cite{pixelsplat,tang2024lgm,mvsplat,gslrm}.
In particular, Long-LRM~\citep{llrm} achieves scene-level GS reconstruction from long-sequence input in a fully feed-forward manner. 
However, these methods mainly focus on color rendering from novel views, whereas the primary goal of MVS is to reconstruct geometry. Without explicit depth supervision, Gaussian-based methods often suffer from the ``floater" issue, where translucent Gaussian clouds appear at incorrect locations. Our method could serve as a strong initialization for future online, feed-forward, large-scene Gaussian reconstructions.

%% file: sec/3_method.tex
\section{Method}


We propose PointRecon, an online, point-based MVS method that incrementally builds and updates a global 3D point cloud from a sequence of posed monocular RGB images.

The core idea of our approach is to represent the scene using a global, feature-augmented point cloud $\mathbf{Q} = \{\mathbf{q}_1,\mathbf{q}_2,\dots\}$, which is continuously updated with new observations. Each point $\mathbf{q}_i = (p_i, f_i, r_i, z_i, \sigma_i, c_i)$ consists of 3D position $p_i \in \mathbb{R}^3$, feature vector $f_i \in \mathbb{R}^C$, initial ray direction $r_i \in \mathbb{R}^3$ (unit vector), distance to the originating camera $z_i \in \mathbb{R}^+$, Gaussian standard deviation $\sigma_i \in \mathbb{R}^+$ of the position along $r_i$, and confidence score $c_i \in \mathbb{R}$. Points can move along their original camera ray based on new information from incoming images $\mathbf{I}_t \in \mathbb{R}^{H \times W \times 3}$ at time $t$. In the following, we use $p^\circ$ for 2D positions and $p$ for their corresponding 3D positions. Furthermore, we use $r_i$ to denote the unit direction vector of a camera ray, while $\mathcal{R}_i$ represents the set of all points lying along that ray, to avoid any confusion in notation.

\begin{figure*}[h]
    \centering
    \includegraphics[width=0.8\linewidth]{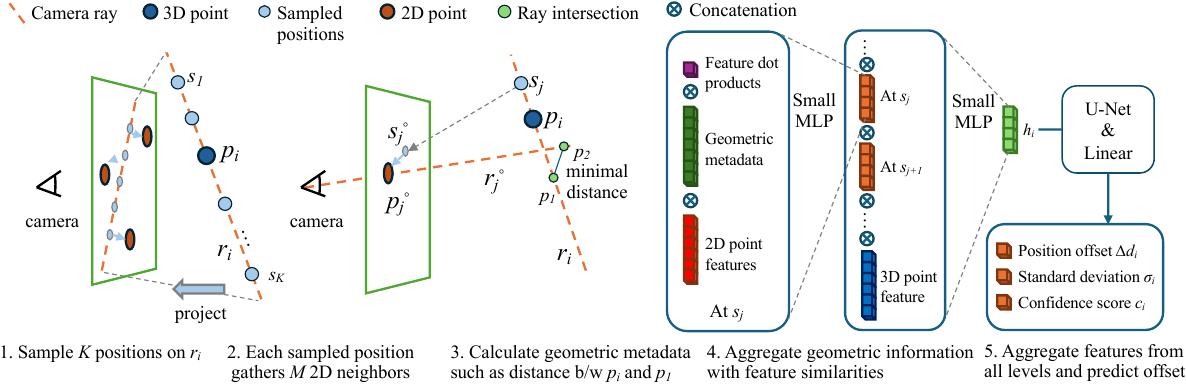}
    \vskip -0.1in
    \caption{\small \textbf{Illustration of the scene update step} (single level shown for simplicity). For a point $p_i$ in the point cloud, we uniformly sample $K$ positions along its ray and project them onto the image plane. Each projected sampled position selects $M$ nearest neighbors in the 2D feature map. 
    We then compute the feature dot product between $p_i$ and each neighbor, along with geometric metadata, such as the distance between $p_i$ and the crossing of their camera rays. The 3D point uses both geometric information and feature similarities to determine its position adjustment along the ray.}
    \label{fig:adjust}
    \vskip -0.15in
\end{figure*}
\begin{figure*}[th]
    \centering
    \includegraphics[width=0.8\linewidth]{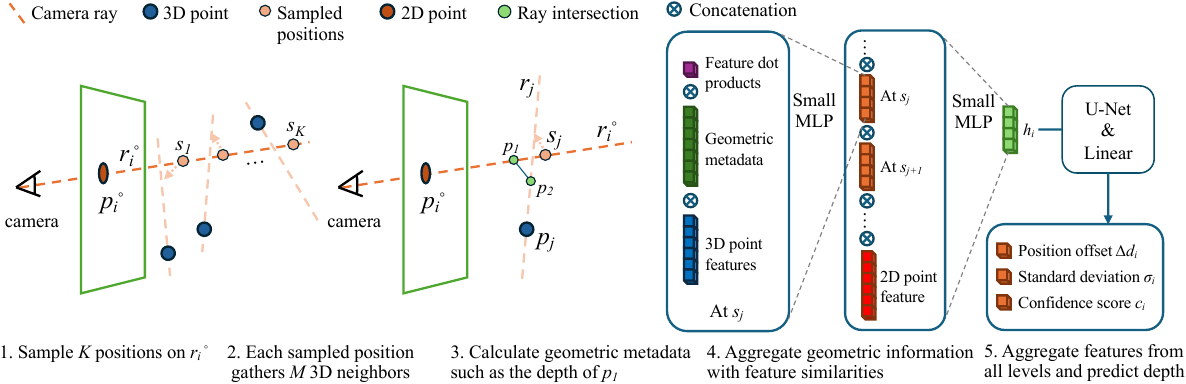}
    \caption{\small \textbf{Illustration of the depth prediction step} (single level shown for simplicity). For a 2D feature point $p_i^\circ$ on the image plane, we uniformly sample $K$ positions along its camera ray. Each sampled position identifies $M$ neighboring points in the point cloud by finding the nearest rays. For each neighbor, we compute its feature dot product with $p_i^\circ$, along with geometric metadata such as the depth at the crossing of the rays. The 2D point uses both geometric data and feature similarities to predict its depth value.}
    \label{fig:match}
    \vskip -0.15in
\end{figure*}

Our key contribution is a ray-based feature matching strategy between 3D points and 2D image pixels, which is robust to errors in the initial 3D positions. This matching is performed in two steps: 1) For each 3D point within the camera frustum of a new image, we project its camera ray onto the image plane to form an epipolar line and gather 2D neighbors near this line (Fig.~\ref{fig:adjust}). These 2D neighbors are used to update the point's feature and refine its 3D position (Sec.~\ref{sec:sim}). 2) For each 2D point in the image, we uniformly sample positions along its camera ray and gather 3D points with the nearest camera rays to form a neighborhood (Fig.~\ref{fig:match}). Feature matching between the 2D point and the 3D neighbors is performed for depth prediction (Sec.~\ref{sec:ism}). The 2D points are then back-projected into 3D space based on their predicted depth.
Finally, we merge the new 3D points with the existing point cloud, removing redundant points by comparing confidence scores of points that project onto the same pixel and keeping only the most confident ones  (Sec.~\ref{sec:merge}).


\subsection{Backbone, Feature Pyramid and Monocular Depth Prediction}\label{sec:mono}


We first pass the image $\mathbf{I}_t$ through an image encoder to produce a set of multi-level feature maps $\{\mathbf{F}^l\}_{l=1}^4$, where level 1 is the coarsest and level 4 the finest. While our approach is compatible with any image encoder, such as ResNets~\citep{resnet}, we specifically use AutoFocusFormer~\citep{autofocusformer}, a transformer-based encoder that excels at preserving key details by automatically locating important points during downsampling. This results in non-uniform 2D point cloud feature maps, represented as $\mathbf{F}^l = \{(p^\circ_1, f_1), (p^\circ_2, f_2), \dots\}$ at each level.


Next, we enhance the feature maps $\{\mathbf{F}^l\}_{l=1}^4$ by constructing a feature pyramid, propagating information from coarser to finer levels. For each level $l \in \{2,3,4\}$, we update $\mathbf{F}^l$ by attending to all coarser levels $\mathbf{F}^k$ (where $k \leq l$) using a transformer block (attention + MLP), starting from the coarsest and moving to the finest.


For the first image in the stream, we perform monocular depth estimation. A linear layer predicts a positive depth value from each point's feature vector $f_i$, along with a positive standard deviation $\sigma_i$ and a confidence score $c_i$. We then back-project the 2D features into 3D using the camera intrinsics and pose, forming the initial multi-level point cloud $\{\mathbf{Q}^l\}_{l=1}^4$. The confidence score $c_i$, a logit in $\mathbb{R}$, is used later during point cloud merging (Sec.~\ref{sec:merge}). While monocular depth estimation is prone to ambiguity, our method quickly corrects errors through scene update.

\vspace{0.2cm}

\subsection{Scene Update}\label{sec:sim}


We begin with a multi-level point cloud $\{\mathbf{Q}^l\}_{l=1}^4$ that may contain position errors. After computing the multi-level feature maps $\{\mathbf{F}^l_t\}_{l=1}^4$ for a new image $\mathbf{I}_t$, we update the visible portion of the point cloud. To do this, we project the 3D camera rays onto the image plane of $\mathbf{I}_t$, creating epipolar lines, and match the 3D point features with the 2D features near these lines. Specifically, for each 3D scene point $\mathbf{q}_i$, we uniformly sample $K$ positions along its camera ray, centered at its 3D location $p_i$. These positions are then projected onto the image plane of $\mathbf{I}_t$, with each finding $M$ nearest 2D features from $\mathbf{F}$. In total, this process collects $KM$ 2D feature points along each ray.


To reduce the memory cost of feature matching, we first reduce the channel size of both the 2D and 3D features from $C$ to 32 using a linear layer across all levels. Then, we compute the dot product between the 3D and the neighboring 2D features. However, for the model to determine the direction in which a point should move, it’s not enough to just compute feature similarities with the neighbors; the model also needs to know \textit{where} the neighbors are, especially given the non-uniform nature of point clouds. To address this, we append geometric metadata to the feature similarity.

The general idea is that if a 3D point $\mathbf{q}_i$ matches with a 2D image point $p_j^\circ$, we should move $\mathbf{q}_i$ to the intersection of its camera ray $r_i$ and the camera ray $r_j^\circ$ of $p_j^\circ$ (the ray from the current camera shooting through $p_j^\circ$) (see Fig.~\ref{fig:adjust}). To enable this adjustment, we gather metadata about the ray intersection. Specifically, we define the intersection as the pair of points—one on each ray—connected by the shortest distance between $r_i$ and $r_j^\circ$:
\begin{align}\small
    p_1(r_i,r_j^\circ) = \arg \min_{p \in \mathcal{R}_i} \min_{p' \in \mathcal{R}_j^\circ} \|p - p'\|\\
    p_2(r_i,r_j^\circ) = \arg \min_{p' \in \mathcal{R}_j^\circ} \min_{p \in \mathcal{R}_i}\|p - p'\|
\end{align}
 , where $p_1$ lies on the ray $r_i$ and $p_2$ lies on the ray $r_j^\circ$. Additionally, we include other metadata such as the ray angle and point distance to further support the reliability of this matching. 
 Concretely, for a 3D point $\mathbf{q}_i$ with camera ray $r_i$, a sampled position $s_j$ on $r_i$, and a 2D neighbor $p_j^\circ$ with camera ray $r_j^\circ$, the list of geometric metadata includes:
\begin{itemize}[noitemsep,topsep=0pt,parsep=0pt,partopsep=0pt]
    \item the signed distance between $p_i$ (the 3D location of $\mathbf{q}_i$) and $p_1$: $(p_1-p_i)\cdot r_i$
    \item the signed distance between $p_i$ and $s_j$: $(s_j-p_i)\cdot r_i$
    \item the signed distance between $s_j$ and $p_1$: $(p_1-s_j)\cdot r_i$
    \item the distance between $p_1$ and the originating camera of $p_i$: $\|p_1-(p_i-z_i\cdot r_i)\|$
    \item the depth of $p_1$ with respect to the current camera
    \item the distance between the two rays: $\|p_1 - p_2\|$
    \item the cosine angle between $r_i$ and $r_j^\circ$: $r_i\cdot r_j^\circ$
    \item the distance between the 2D projection of $s_j$ and $p_j^\circ$: $\|s_j^\circ - p_j^\circ \|$
    \item the probability density at $p_1$ within the Gaussian distribution centered at $p_i$ with standard deviation $\sigma_i$: $\frac{1}{\sigma_i\sqrt{2\pi}}\exp(-\frac{1}{2}(\frac{p_1-p_i}{\sigma_i})^2)$
    \item a binary mask indicating whether $s_j$ lies inside the current camera frustum
\end{itemize}

We concatenate the dot products, metadata, and reduced 2D features at each sampled position and pass them through a small MLP. Next, we concatenate the output vectors from all sampled positions along with the reduced feature of $\mathbf{q}_i$, feeding them through another small MLP. The final output vector $h_i$ represents the information that $\mathbf{q}_i$ gains from the image $\mathbf{I}_t$. 

Finally, we use a small U-Net to aggregate features across all levels. During the encoding process, each level first attends to the features one level finer and then to itself. In the decoding process, each level first attends to the features one level coarser and then to itself. From the updated feature $h_i$, we use a linear layer to predict a position offset $\Delta d_i$ along the ray, as well as an updated standard deviation $\sigma_i$ and a confidence score $c_i$. At levels other than the coarsest, we leverage point movements from the coarser point cloud to guide the offset predictions. Specifically, each point at these levels collects the predicted offsets from nearby points in the coarser point cloud, projects these offsets onto its own ray, interpolates them using inverse distance weights, and appends the result to $h_i$ before making the final offset prediction.
Additionally, we append ${h_i}$ to the original point features ${f_i}$ and pass them through a linear layer to produce the final updated features for the scene.


\vspace{0.2cm}

\subsection{Depth Prediction}\label{sec:ism}

Next, we predict the depth values for the 2D points. This feature matching process closely mirrors the one used in scene update. For each 2D point $p^\circ_i$, let $r_i^\circ$ represent the ray passing through it from the current camera view. We begin by predicting a rough depth value for $p^\circ_i$ using the monocular depth prediction module. Then, we uniformly sample $K$ positions along $r_i^\circ$ within a range centered around the predicted 3D position $p_i'$. 
Each sampled position identifies the $M$ nearest rays from the scene, resulting in a collection of $KM$ 3D neighbors along the ray $r_i^\circ$. 

We opt to find the nearest rays, rather than the nearest points, because the point positions are predicted and might be inaccurate. Additionally, by first sampling positions along $r_i^\circ$ we avoid concentrating matches at shallow depths. This occurs because ray intersections from neighboring cameras tend to cluster more densely at shallow depths, which would skew the matching process if we directly searched for rays closest to $r_i^\circ$.

We again compute the feature dot product between the reduced 2D and 3D features. As in the scene update step, we append geometric metadata to the dot products. The core idea is that if the 2D point $p^\circ_i$ matches with a 3D ray $r_j$, then the depth of the crossing between $r_i^\circ$ and $r_j$ is likely the correct depth for $p^\circ_i$.
Concretely, for a 2D point $p_i^\circ$ with camera ray $r_i^\circ$, a sampled position $s_j$ on $r_i^\circ$, and a 3D neighbor with position $p_j$ and ray $r_j$, the list of geometric metadata includes:
\begin{itemize}[noitemsep,topsep=0pt,parsep=0pt,partopsep=0pt]
    \item the depth of the crossing $p_1(r_i^\circ, r_j)$ with respect to the current camera
    \item the depth of the sampled position $s_j$
    \item the depth difference between $p_1$ and the sampled position $s_j$
    \item the distance between $s_j$ and $r_j$: $\min_{p\in \mathcal{R}_j}\|p-s_j\|$
    \item the distance between the two rays: $\|p_1 - p_2\|$
    \item the distance between $p_1$ and $p_j$: $\|p_j-p_1\|$
    \item the distance between $s_j$ and $p_j$: $\|p_j-s_j\|$
    \item the distance between $p_1$ and the current camera
    \item the distance between $p_2$ and the originating camera of $p_j$: $\|p_2-(p_j-z_j\cdot r_j)\|$
    \item the cosine ray angle between $r_i^\circ$ and $r_j$: $r_i^\circ\cdot r_j$
    \item the cosine ray angle between $r_i^\circ$ and the ray connecting $s_j$ and the originating camera of $p_j$: $r_i^\circ \cdot \frac{(s_j-(p_j-z_j\cdot r_j))}{\|(s_j-(p_j-z_j\cdot r_j))\|}$
    \item the probability density at $p_2$ within the Gaussian distribution centered at $p_j$ with standard deviation $\sigma_j$: $\frac{1}{\sigma_j\sqrt{2\pi}}\exp(-\frac{1}{2}(\frac{p_2-p_j}{\sigma_j})^2)$
    \item a binary mask indicating whether $p_j$ lies in front of the current camera
\end{itemize}

Similar to the scene update step, we concatenate the metadata with the reduced 3D features and the feature dot products at each sampled position, passing them through a small MLP. The outputs from all sampled positions are then combined with the reduced feature of $p^\circ_i$, and this combined vector is processed by another small MLP to produce $h_i$, which represents the information $p^\circ_i$ has gained from the scene point cloud.

Finally, we apply a small U-Net to aggregate features across all levels, similar to the scene update step. From the updated features $h_i$, we directly predict the depth values $d_i$, the standard deviation $\sigma_i$ and the confidence score $c_i$ using a linear layer.
At levels other than the coarsest, we utilize the depth predictions from the coarser levels for guidance. Each point at these levels gathers predicted depths from nearby points in the coarser feature map, interpolates them using inverse distance weighting, and appends the result to $h_i$ before making the final depth prediction.
Additionally, We append $h_i$ to the original 2D point features $f_i$, feeding them through a linear layer to generate the final updated features. The 2D points are then back-projected into 3D space using the predicted depth, and this new set of points is added to the existing scene point cloud.

\subsection{Point Cloud Merging}\label{sec:merge}

While the overlap in the source images is necessary for stereo matching, it also causes redundancy in the point cloud. When a 2D point successfully matches with a 3D point, both represent the same surface area, while ideally, we would like to keep only one of them. This leads to our point cloud merging step.


First, we find all the cameras in the history whose camera frustum has overlap with the current camera frustum. Then, we project the points in the current camera frustum onto the image planes of these cameras.
During training, to train $c_i$, we first render the depth value at each pixel $p_i^\circ$ by interpolating the depth of the points with the $c_i$:
\begin{equation}
\small
d_i = \sum_{P(\mathbf{q}_j, I_t) = p_i^\circ} \text{softmax}(c_j)d_j 
\end{equation}
where $P(\mathbf{q}_j, \mathbf{I}_t)$ projects $\mathbf{q}_j$ to the image plane of $\mathbf{I}_t$.
Thus, for the model to render the correct depth, it needs to boost the $c_i$ of a point on the true surface and reduce $c_i$ of those points at incorrect location. Then, we trim the point cloud by only keeping the point with the highest confidence score at each pixel. 
During inference, we keep the point with minimal depth to the camera at each pixel $p_i^\circ$:
\begin{equation}
\small
d_i = \min_{\begin{subarray}{c}P(\mathbf{q}_j, \mathbf{I}_t) = p_i^\circ\\c_j>\epsilon\end{subarray}} d_j 
\end{equation}
where $\epsilon$ is empirically set to 0. The value $d_i$ is also used as the rendered depth value from the point cloud during evaluation.

Note that while we train the model using 9-view local windows, our goal is to seamlessly reconstruct the entire scene during evaluation. As the sequence length increases, the number of cameras sharing a partial view can reach into the hundreds, leading to a significant computational burden. To address this, we employ a heuristic: if a camera is not among the $K$ most recent cameras in the current merging step, we retain all points covering the pixels of that camera, preventing their deletion, and remove the camera from the current merging algorithm. We empirically set $K=16$.


\subsection{Loss Functions}
We train our model by supervising the depth maps produced by different modules of our model: the monocular depth prediction, the depth prediction with feature matching and the depth maps rendered at the point merging step. The three types of depth maps use the same set of losses: an $L1$ depth loss, a gradient loss and a normal loss following ~\citep{simplerecon}.

We apply the $L1$ depth loss 
\begin{equation} \small \label{eq:loss_dep}
    \mathcal{L}_{\text{depth}} = \dfrac{1}{HW}\sum_{l=1}^4 \dfrac{1}{l^2}\sum_i |\log d_{i 
    }^l - \log d_i^\text{gt}|
\end{equation}
to all four levels. We upsample the depth values from the coarse levels to full resolution using inverse distance weighted interpolation. We apply the gradient loss
\begin{equation}\label{eq:loss_grad}\small
    \mathcal{L}_{\text{grad}} = \dfrac{1}{HW}\sum_{r=0}^3 \sum_i|\nabla d_{i \downarrow_{2^r\times}
    }^4 - \nabla d_{i \downarrow_{2^r\times}}^\text{gt}|
\end{equation}
to only the finest level, but both to the full resolution and to the three downsampled versions (we downsample the resolution by half each time). $\nabla$ represents the first-order spatial gradients, and $\downarrow_{2^r\times}$ means downsampling by a rate of $2^r$. We also apply the normal loss
\begin{equation}\label{eq:loss_normal}\small
    \mathcal{L}_{\text{normal}} = \dfrac{1}{2HW}\sum_i 1-\mathbf{N}_{i}\cdot \mathbf{N}_i^\text{gt}
\end{equation}
to the finest level, where $\mathbf{N}_i$ is the normal vector calculated from predicted depth and camera intrinsics. Additionally, to supervise the scene update step, we calculate the difference between the distance from each point to its camera and the ground-truth value:
\begin{equation}
    \label{eq:loss_adj}\small
    \mathcal{L}_{\text{update}} = \dfrac{1}{N}\sum_{l=1}^4 |z_i - z_i^\text{gt}|
\end{equation}
where $z_i$ is the distance to the camera computed from the predicted 3D position $p_i$ and $z_i^{gt}$ is the ground truth distance. $N$ is the total number of points from all 4 levels.
Overall, our total loss is
\begin{equation}\label{eq:total}
    \small 
    \mathcal{L}_{\text{total}} = \mathcal{L}_{\text{depth}} + 
    \mathcal{L}_{\text{grad}} +
    \mathcal{L}_{\text{normal}} +
    \mathcal{L}_{\text{update}}
\end{equation}


%% file: sec/4_experiments.tex
\begin{table*}[h]
\begin{center}
\begin{scriptsize}
\renewcommand{\arraystretch}{1.2}
\begin{tabular}{@{}$c^l@{ }^c@{ }^c@{ }@{ }^c@{ }^c@{ }^c@{ }^c@{ }^c@{ }^c@{}}
\toprule
\rowstyle{\bfseries}
\makecell{Recon\\Type} &Method & \makecell{Non-\\Volumetric} & \makecell{Latency\\(ms/frame)} & Abs Diff$\downarrow$ & Abs Rel$\downarrow$ & Sq Rel$\downarrow$ & $\delta\!<\!1.05$$\uparrow$ & $\delta\!<\!1.25$$\uparrow$ & Comp$\uparrow$ \\
\midrule
\multirow{5}{*}{Offline} & COLMAP~\cite{colmap} & \cmark & / & 0.264 & 0.137 &  0.138 & - & 83.4 & 87.1 \\
&Atlas~\cite{atlas} & \xmark & / &  0.123 & 0.065 &  0.045 &- & 93.6 & \textbf{99.9} \\
&VoRTX~\cite{vortx} & \xmark &  / & 0.092 & 0.058 & 0.036 &- & 93.8 & 95.0 \\
&CVRecon~\cite{cvrecon}& \xmark &  / & 0.078 & 0.047 & 0.028 & -  & 96.3 & -\\
&FineRecon~\cite{finerecon} & \xmark & / &  \textbf{0.069} & \textbf{0.042} & \textbf{0.026} & \textbf{86.6} & \textbf{97.1} & 97.2\\
\midrule

\multirow{5}{*}{Online} & NeuralRecon~\cite{neuralrecon} & \xmark &  90 & 0.106 & 0.065 &  0.031 &- & 94.8 & 90.9 \\
&TF~\cite{transformerfusion} & \xmark & 326 &  0.099 & 0.065 &  0.042 & - & 93.4 &  90.5 \\

&SimpleRecon~\cite{simplerecon} & \cmark & 72 &  \textbf{0.083} & \textbf{0.046} &  \textbf{0.022} & - & 95.4 &  94.4 \\
\cmidrule(lr){2-10}
&PointRecon (4cm) & \cmark &  618 &  \textbf{0.085} & 0.054 & \textbf{0.022} & 71.9 & \textbf{96.4} & \textbf{94.8} \\
&PointRecon (2cm) & \cmark &  618 &  0.087 & 0.055 & 0.024 & \textbf{72.1} & \textbf{96.2} & \textbf{94.6} \\
\bottomrule
\end{tabular}
\end{scriptsize}
\end{center}
\vspace{-0.6cm}
\caption{\small \textbf{Depth Map Evaluation.} Depth map quality for the ScanNetv2 test split, rendered from the reconstructed 3D meshes. We follow the evaluation protocol of Atlas~\citep{atlas}. \textbf{-} indicates that the metric was not reported in the original paper. Offline methods assume access to the entire sequence and do not support local updates. The numbers in brackets indicate the TSDF Fusion resolution. While we present results for both 2cm and 4cm resolutions (with 4cm commonly used in volumetric methods), our point-based representation is not constrained by any specific TSDF Fusion resolution.}
\label{tb:depth}
\vskip -0.15in
\end{table*}

\begin{table}[h]
\begin{center}
\begin{scriptsize}
\renewcommand{\arraystretch}{1.2}
\begin{tabular}{@{}$c^l@{}^c@{}^c@{}^c@{}^c@{}^c@{}^c@{}}
\toprule
\rowstyle{\bfseries}
\makecell{Recon\\Type} & Method &  Comp$\downarrow$ & Acc$\downarrow$ & Chamfer$\downarrow$ & Prec$\uparrow$ & Recall$\uparrow$ & F-Score$\uparrow$ \\
\midrule
\multirow{4}{*}{Offline} & COLMAP
& \textbf{0.069} & 0.135 &  0.102 & 0.634 & 0.505 & 0.558 \\
&Atlas
& 0.084 & 0.102 &  0.093 & 0.598 & 0.565 & 0.578 \\
&VoRTX
& 0.082 & 0.062 & 0.072 & 0.688 & 0.607 & 0.644 \\
& CVRecon
&  0.077 
& \textbf{0.045} & \textbf{0.061} & \textbf{0.753} & \textbf{0.639} & \textbf{0.690} \\
\midrule

\multirow{7}{*}{Online} & DeepVMVS
& 0.076 & 0.117 &  0.097 & 0.451 & 0.558 & 0.496 \\

& NeuralRecon
& 0.128 & \textbf{0.054} & 0.091 & 0.684 & 0.479 & 0.562 \\
&TF
& 0.099 & 0.078 &  0.089 &  0.648 &  0.547 &  0.591 \\

&SimpleRecon
& 0.078 & 0.065 &  \textbf{0.072} &  0.641 &  0.581 &  \textbf{0.608} \\
&VisFusion
& 0.105 & \textbf{0.055} & 0.080 & \textbf{0.695} & 0.527 & 0.598 \\
\cmidrule(lr){2-8}
&PointRecon (4cm) 
& 0.082 & 0.086 & 0.084 & 0.568 & 0.556 & 0.560 \\
&PointRecon (2cm)
& \textbf{0.073} & 0.083 & 0.078 & 0.586 & \textbf{0.612} & 0.599  \\
\bottomrule
\end{tabular}
\end{scriptsize}
\end{center}
\vspace{-0.6cm}
\caption{\small \textbf{Mesh Evaluation.} Mesh reconstruction quality for the ScanNetv2 test split, following the evaluation protocol of Atlas~\citep{atlas}. The number in parentheses represents the TSDF Fusion resolution.}
\label{tb:mesh}
\vskip -0.15in
\end{table}


\section{Experiments}

\subsection{Implementation Details}
We select AFF-Mini, the smallest variant from \cite{autofocusformer} consisting of only 6.75M parameters, as our image encoder backbone. Our model is trained and evaluated on the ScanNetv2~\citep{scannet} dataset. We resize the input resolution to $640\times 480$. The finest-level depth maps are predicted at a resolution of $160\times 120$, creating 19,200 points each new frame.

\textbf{Model hyperparameters.} In both the scene update step and the depth prediction step, we set the sample number $K=32$ and number of neighbors $M=1$. The sampling range is configured to be 1.5m, while the point update offset range is limited to 5m. Following the design of AFF, all transformer blocks in the model perform local attention with a neighborhood size of 48 and an MLP expansion ratio of 2.0 by default. When level $i$ attends to a coarser level $j$, the neighborhood size is reduced to $\max(4, \frac{48}{4^{i-j}})$.

\textbf{Relative position embedding.} Within the 2D local attention layers, we follow AFF by embedding the relative position between two points $p_i^\circ$ and $p_j^\circ$ using the position difference $p_i^\circ-p_j^\circ$, the distance $\|p_i^\circ-p_j^\circ\|$, and the sine and cosine of the angle between $p_i^\circ-p_j^\circ$ and the positive $x$-axis. In the 3D local attention layers, for two points  $p_i$ and $p_j$, we similarly embed their relative position, but replace the 2D angular components with the sine and cosine values of the angle between $p_i-p_j$ and the ground plane.


\textbf{Training details.} We begin with training the model on image pairs for 2 epochs, using the two images to mutually serve as sources for feature matching, without any point merging at this stage. Then, we train the model on 9-view local image windows for 6 epochs, feeding the images sequentially into the model and performing point merging at each time step. We use a learning rate $10^{-4}$ for the first $4$ epochs, reducing it by a factor of 0.1 at the fifth epoch and the eighth epoch. We train with the AdamW optimizer~\citep{adamw} with a weight decay of $10^{-4}$.


During training, we apply random color augmentation to the input RGB images with a probability of 0.8, using TorchVision~\citep{torchvision}. The color augmentation parameters include brightness set to 0.4, contrast to 0.4, saturation to 0.2, and hue to 0.1. Additionally, we randomly reverse the order of the input sequences with a probability of 0.5. For keyframe selection, we follow the methodology outlined in \citep{deepvideomvs}. We reduce the weight for the loss of monocular depth prediction by a factor of 0.5 at the fifth and eighth epochs.

\textbf{Evaluation details.}
During evaluation, we sequentially feed the keyframe images into the model, performing point merging at each time step. Once the point cloud for the entire scene is constructed, we render depth maps using the keyframe cameras by projecting the points onto the image planes. Each pixel in the rendered depth maps corresponds to the depth of the closest point (as detailed in Sec.~\ref{sec:merge}).  To construct a 3D mesh from the rendered depth maps, we apply TSDF fusion~\citep{tsdffusion}.

\begin{figure*}[t]
    \centering
    \begin{subfigure}{0.22\linewidth}
         \includegraphics[width=1.0\columnwidth]{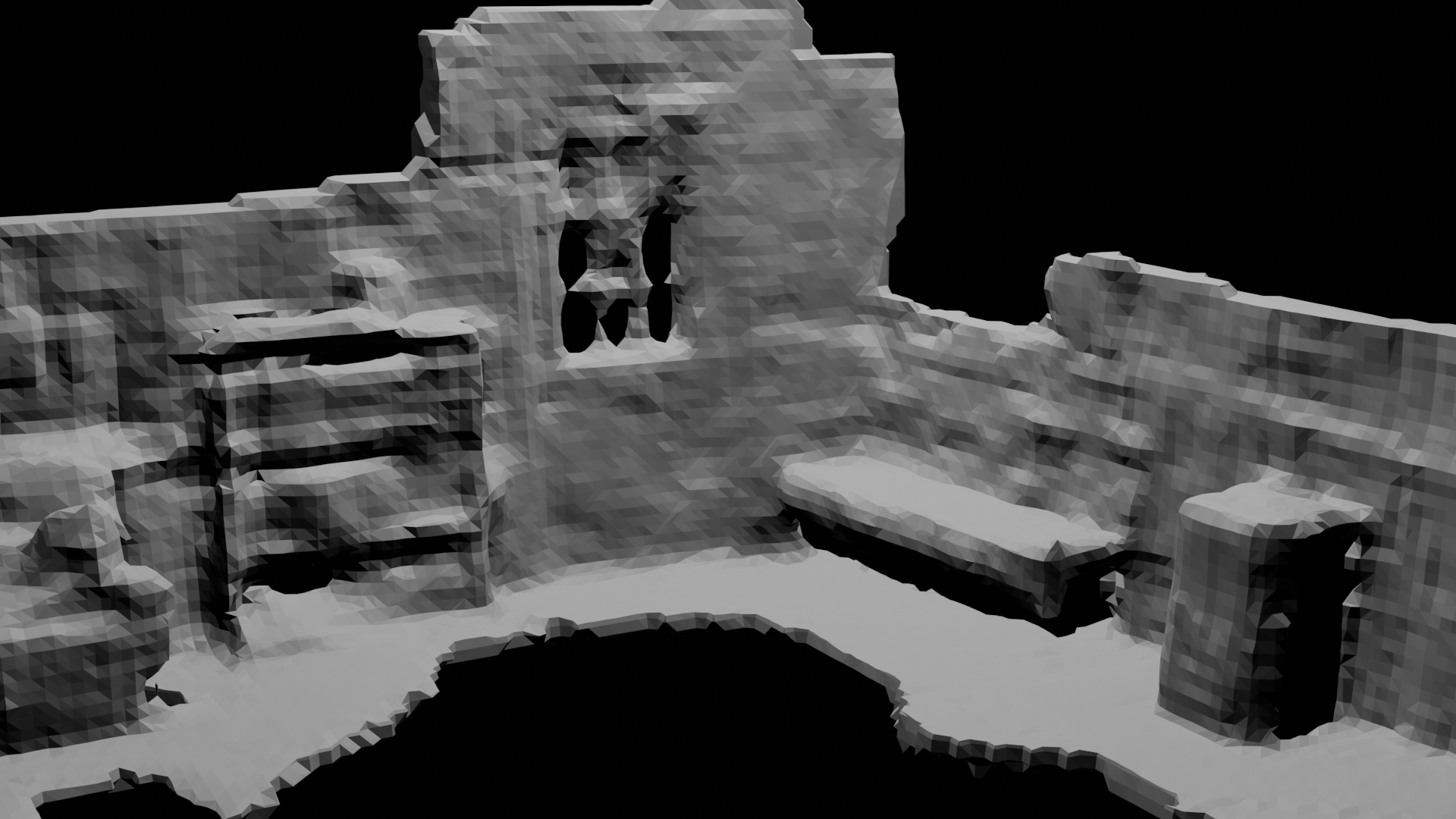} 
      \caption{NeuroRecon}
    \end{subfigure} 
    \begin{subfigure}{0.22\linewidth}
         \includegraphics[width=1.0\columnwidth]{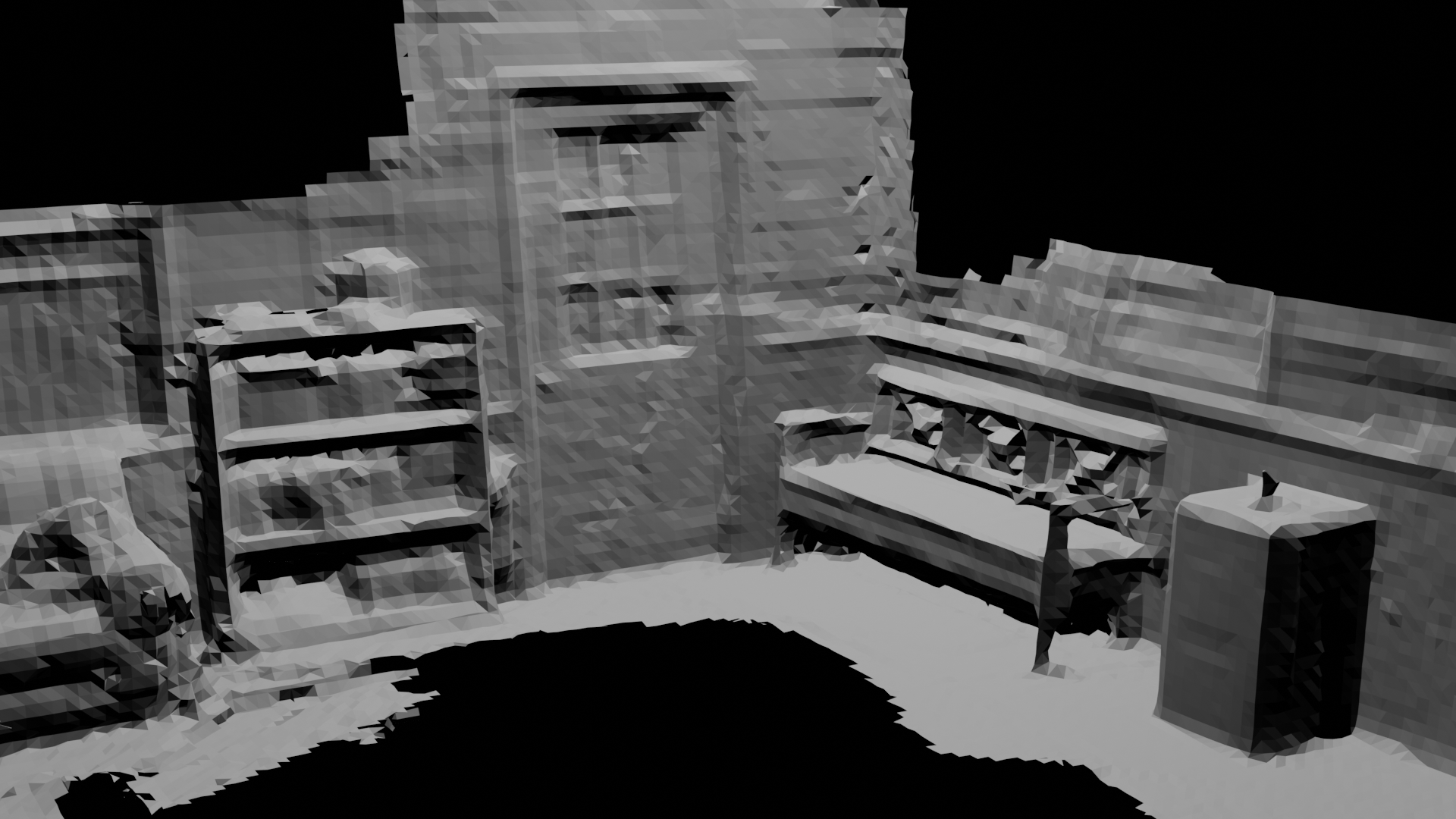} 
      \caption{SimpleRecon}
    \end{subfigure} 
    \begin{subfigure}{0.22\linewidth}
         \includegraphics[width=1.0\columnwidth]{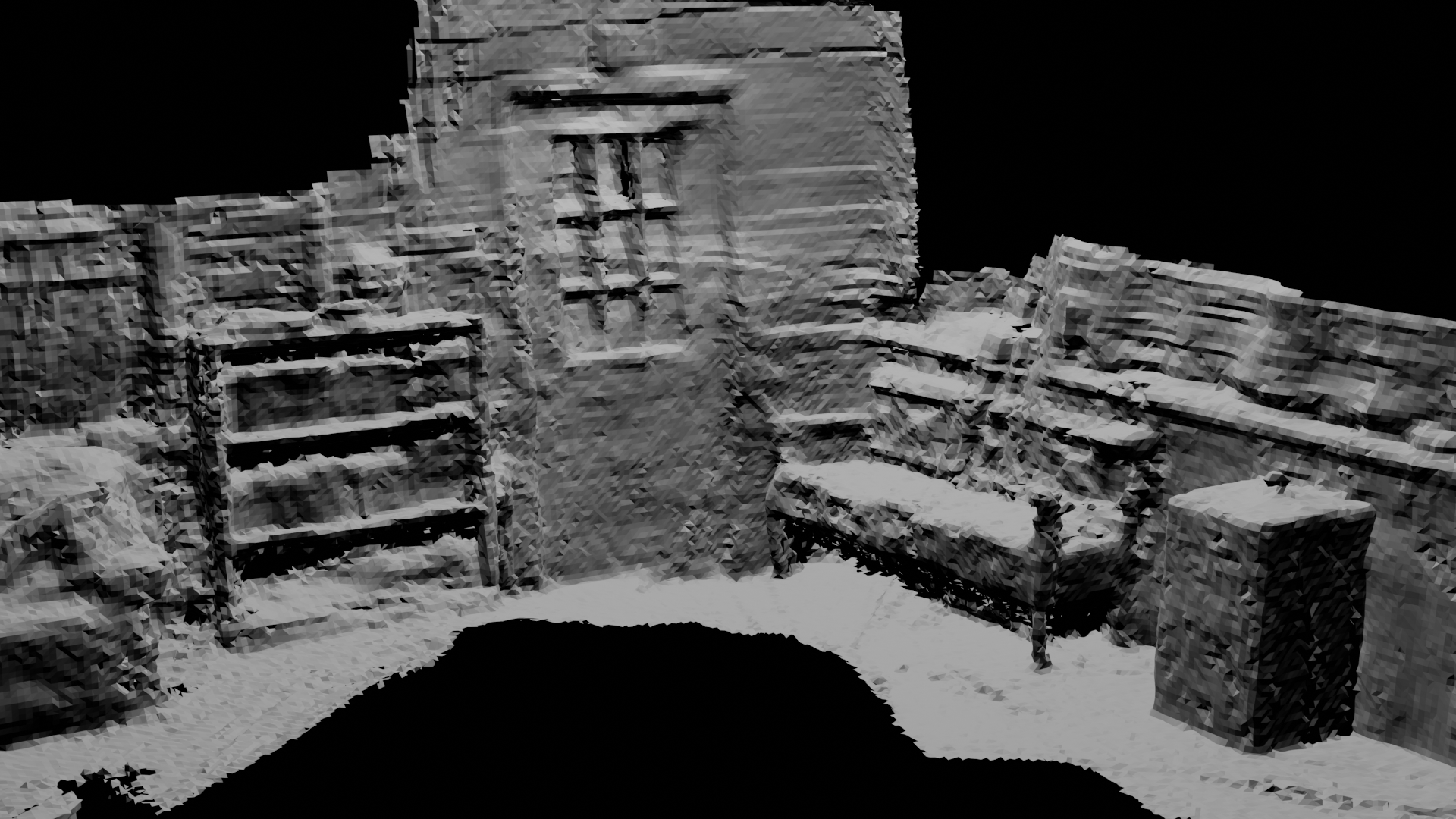} 
      \caption{Ours}
    \end{subfigure} 
    \begin{subfigure}{0.22\linewidth}
         \includegraphics[width=1.0\columnwidth]{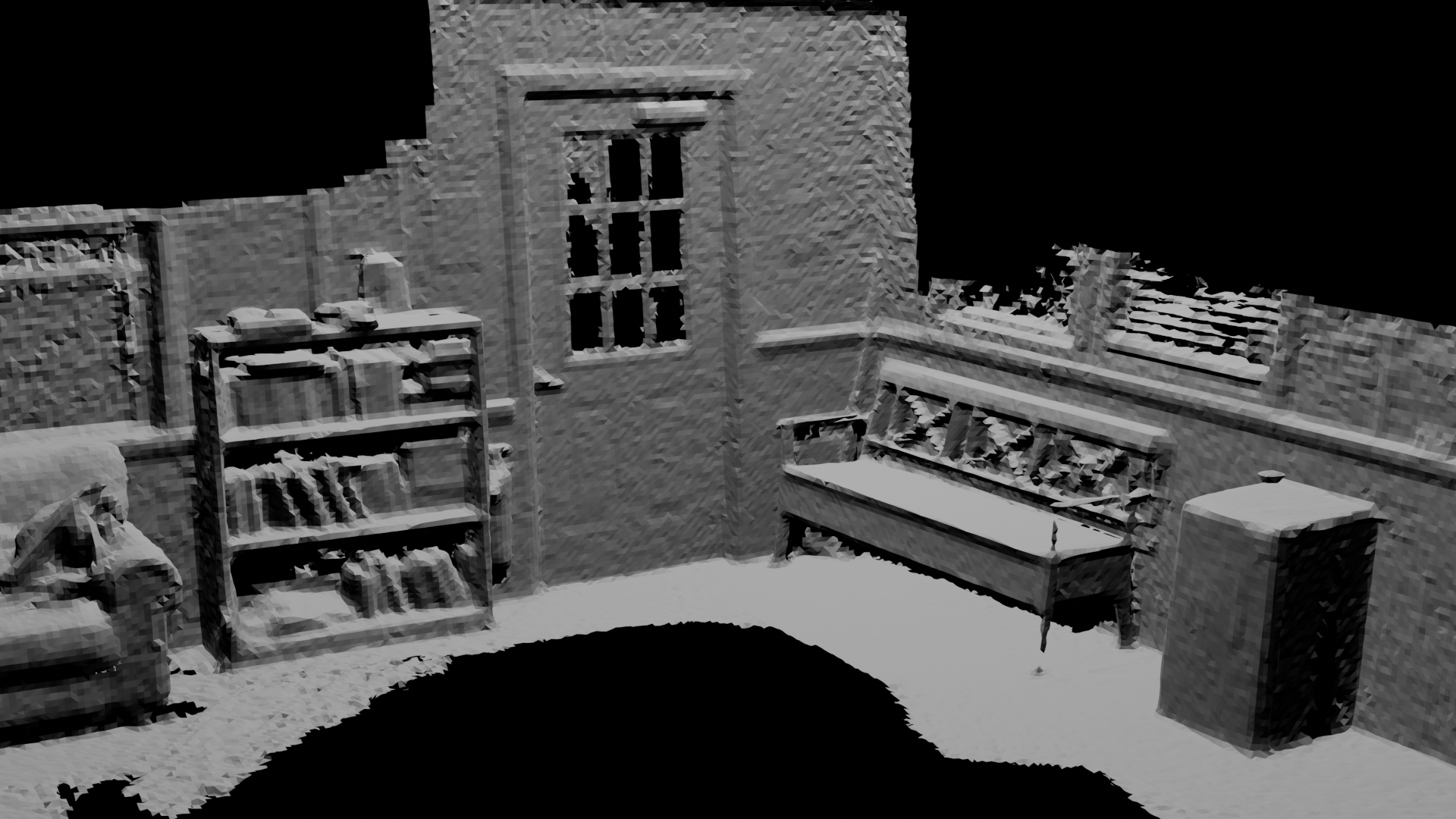} 
      \caption{Ground truth}
    \end{subfigure} \\
    \begin{subfigure}{0.22\linewidth}
         \includegraphics[width=1.0\columnwidth]{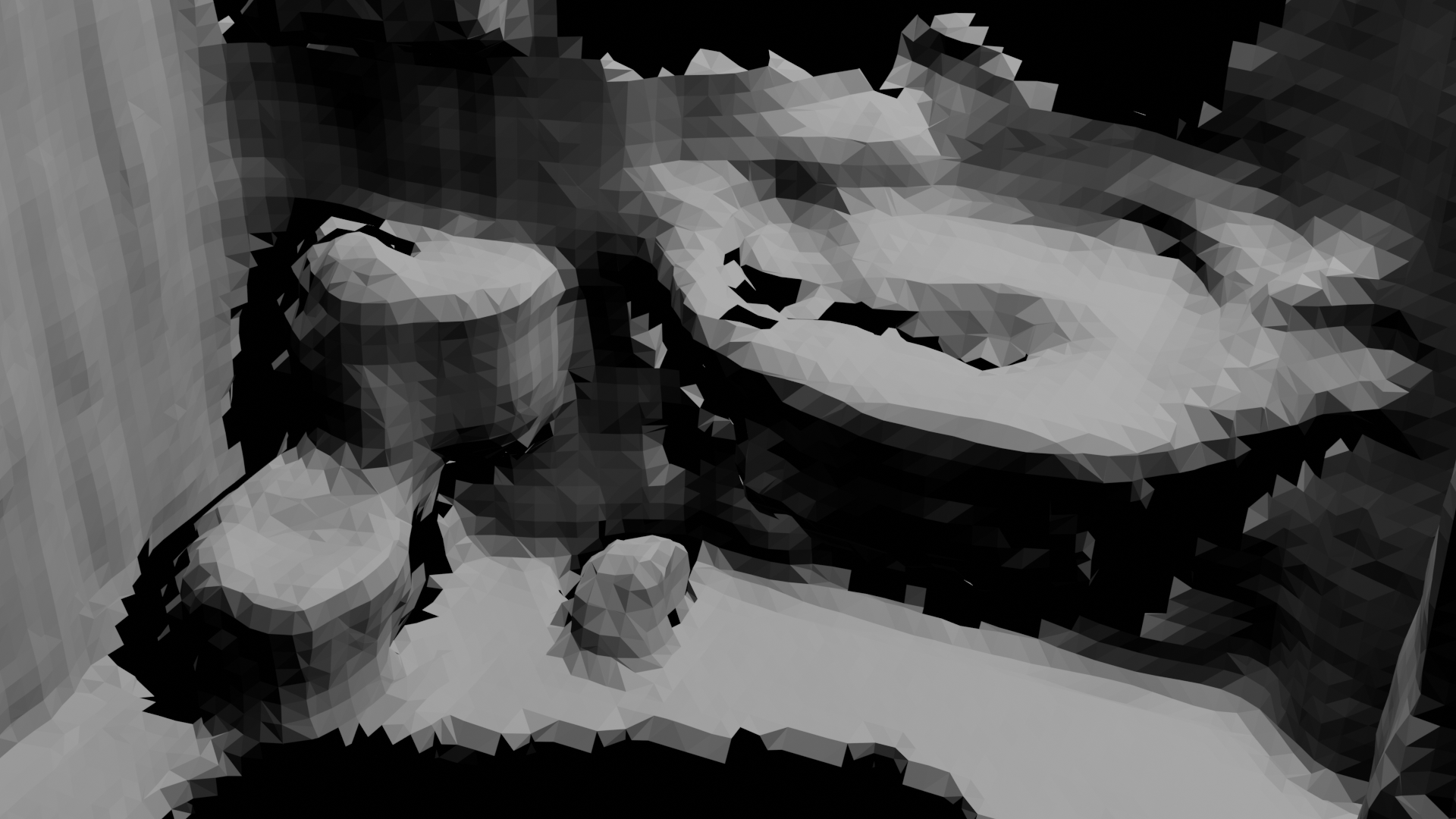} 
      \caption{NeuroRecon}
    \end{subfigure} 
    \begin{subfigure}{0.22\linewidth}
         \includegraphics[width=1.0\columnwidth]{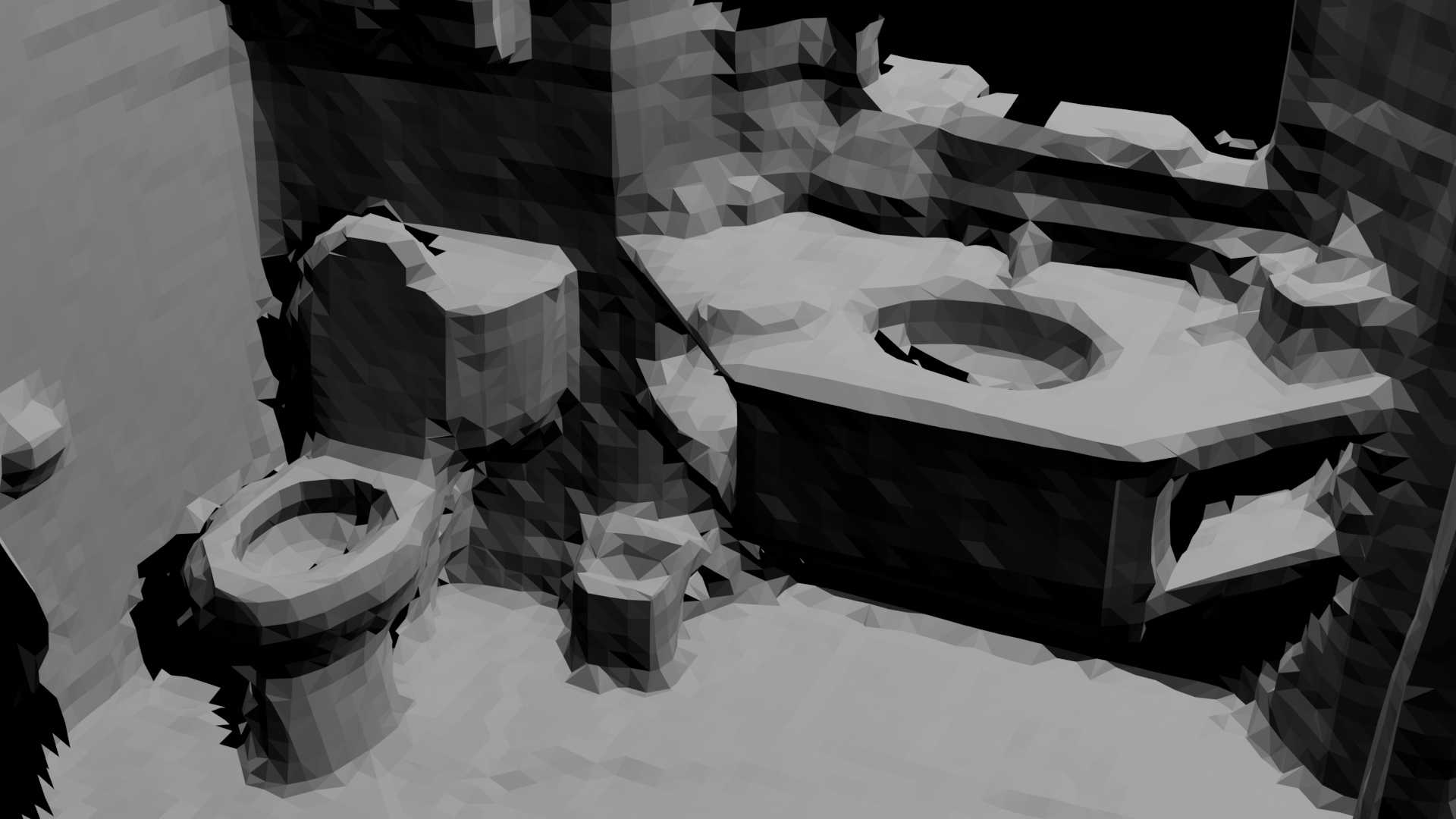} 
      \caption{SimpleRecon}
    \end{subfigure} 
    \begin{subfigure}{0.22\linewidth}
         \includegraphics[width=1.0\columnwidth]{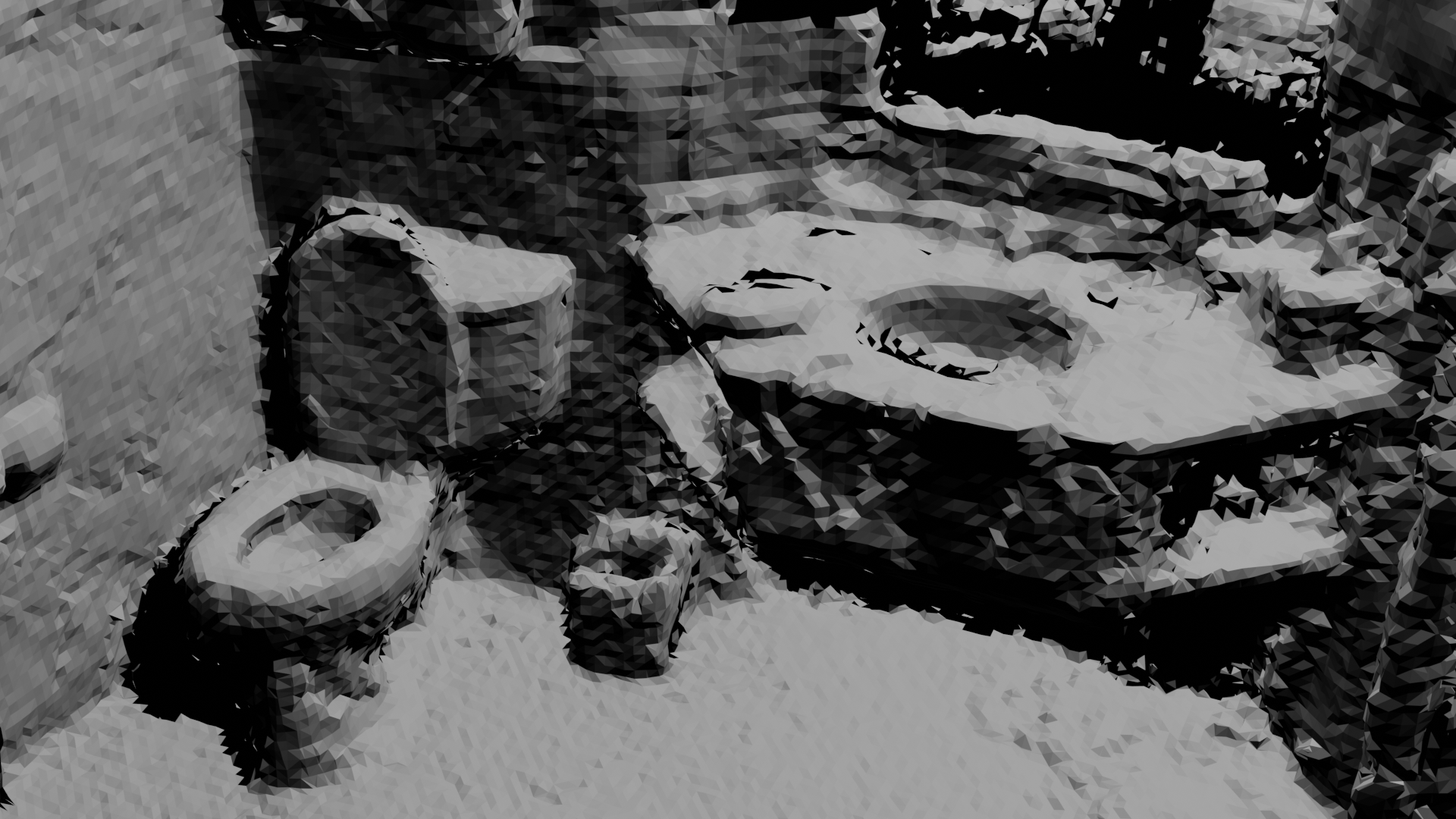} 
      \caption{Ours}
    \end{subfigure} 
    \begin{subfigure}{0.22\linewidth}
         \includegraphics[width=1.0\columnwidth]{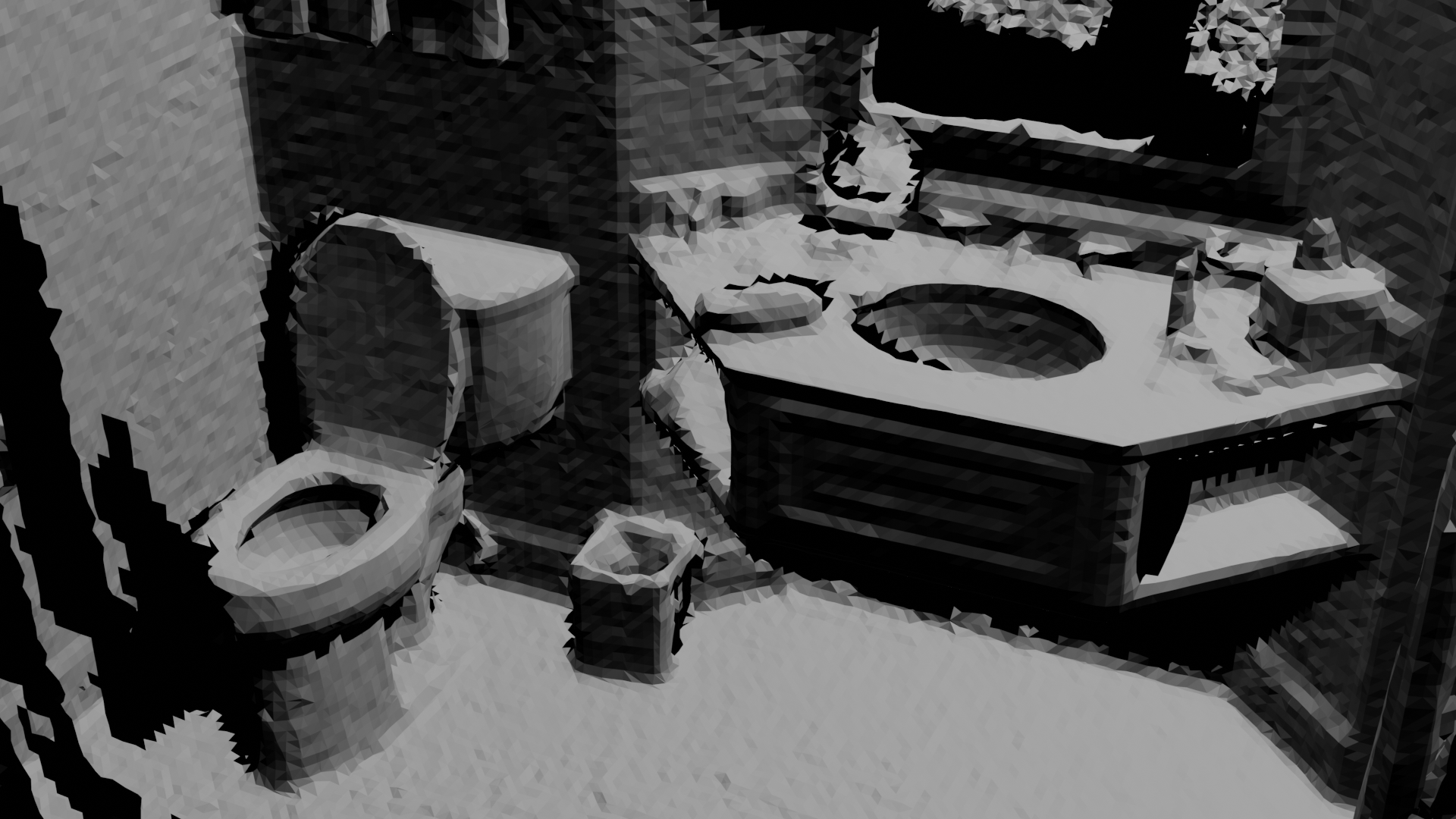} 
      \caption{Ground truth}
    \end{subfigure} \\
    \vskip -0.1in
\caption{\small Visualization of generated meshes (please magnify when viewing). We first render depth maps from the scene point cloud and then fuse them using TSDF fusion to generate mesh. Our method produces more detailed reconstructions compared to previous work, though less smooth surfaces may result from the inherent discrete nature of point clouds.
\textbf{Check out our project page \url{https://arthurhero.github.io/projects/pointrecon/} for incremental reconstruction videos.}
}
\vskip -0.15in
\end{figure*}

\subsection{Experiment Results}

\textbf{ScanNet.} ScanNetv2~\citep{scannet} is an indoor RGB-D video dataset comprising scans of 1,613 rooms, with 1,201 rooms designated for training, 312 for validation, and 100 for testing. Each scan typically includes thousands of image frames, with approximately 1/10 of the frames designated as keyframes. 
We compare our results with previous work in terms of both 2D depth map quality  (Table~\ref{tb:depth}) and 3D mesh quality (Table~\ref{tb:mesh}), where the depth maps are rendered from the reconstructed 3D meshes. To showcase the flexibility of our point representation, we generate meshes at two TSDF fusion resolutions, 2cm and 4cm, demonstrating that our approach is not constrained by TSDF resolution.
As shown in the tables, our model achieves competitive performance among online methods, particularly excelling in depth map metrics such as ``Abs Diff" and ``$\delta<1.25$," and delivering comparable results in Chamfer distance and F-score. Additionally, our method uniquely offers resolution flexibility and viewpoint consistency, attributes not captured by standard evaluation metrics.
A notable strength of our approach is its top-ranking recall score among all online methods, highlighting our model's ability to achieve high coverage of ground-truth surfaces. However, a lower precision score reflects some surface redundancy in the reconstructed meshes. This issue stems from the current merging heuristic, which retains points from earlier camera views. While this strategy ensures robust coverage, it may also preserve outdated or less accurate points, leading to redundant surfaces. Addressing this limitation is a promising direction for future optimization.

\textbf{Profiling.} Following \cite{simplerecon}, we measure the inference speed of our model on a single A100 GPU. On average, the image encoder incurs a latency of 1.8ms per image. The monocular depth predictor requires 8ms, the scene update step takes 215ms, depth prediction with feature matching takes 276ms, and the point merging step adds 118ms. The most computationally intensive component is the transformer U-Net, used in both the scene update and depth prediction steps. The merging process also becomes slower as the point cloud grows.
While point-based representations eliminate grid constraints, they introduce computational challenges, particularly in $K$ nearest neighbor searches, which become slower as the point cloud scales. To enhance efficiency, future work will prioritize simplifying the network architecture and further reducing the point cloud size.
It is important to note that, despite these computational demands, our approach remains an online method capable of continuously updating the scene with new inputs. In contrast, offline methods require access to the entire sequence before reconstruction, learning features from all images jointly, which inherently imposes an upper limit on sequence length.

\begin{table}[h]
\begin{center}
\begin{scriptsize}
\renewcommand{\arraystretch}{1.2}
\begin{tabular}{@{}$c@{ }^l@{}^c@{}^c@{}^c@{}^c@{}^c@{}}
\toprule
\rowstyle{\bfseries}
\makecell{Scene\\Update} & \makecell{Model\\Component}& Abs Diff$\downarrow$ & Abs Rel$\downarrow$ & Sq Rel$\downarrow$ & $\delta<1.05$$\uparrow$ & $\delta<1.25$$\uparrow$ \\
\midrule
\multirow{3}{*}{Off} & Monocular &  0.206 & 0.134 & 0.053 & 29.7 & 82.8 \\
&Feature Matching &  0.130 & 0.083 & 0.028 & 50.7 & 93.3 \\
&Merging &  0.109 & 0.096 & 0.023 & 57.8 & 95.5 \\
\midrule
\multirow{3}{*}{On} & Monocular &  0.206 & 0.134 & 0.053 & 29.7 & 82.8 \\
&Feature Matching &  0.113 & 0.071 & 0.024 & 58.4 & 94.5 \\
&Merging & \textbf{0.085} & \textbf{0.053} & \textbf{0.017} & \textbf{69.9} & \textbf{97.0} \\
\bottomrule
\end{tabular}
\end{scriptsize}
\end{center}
\vspace{-0.6cm}
\caption{\small \textbf{Ablation studies}. Depth map quality on the ScanNet test split key frames produced by different components of our model. The results show that both the Feature Matching and the Merging modules consistently enhance depth prediction quality. The importance of the Scene Update step is demonstrated by the consistent decrease in the depth map quality when it is omitted (Scene Update does not affect the monocular depth prediction as expected).}
\label{tb:ablation}
\vskip -0.15in
\end{table}

\textbf{Ablation.} We analyze the contributions of different components of our model by evaluating the quality of the depth maps produced by these components on the ScanNet test split key frames. To assess the necessity of the scene update step, we train an additional model without this component. As shown in Table~\ref{tb:ablation}, feature matching significantly improves depth prediction accuracy compared to monocular prediction. Point merging further enhances the accuracy by leveraging predictions from multiple cameras, mitigating ambiguities that a single camera might face. The confidence-based merging results in a higher-quality point cloud. The scene update step is also crucial, as it refines the point cloud, leading to more accurate feature matching and merging in subsequent steps.

%% file: sec/5_conclusion.tex
\section{Conclusion}

We propose PointRecon, an online, point-based 3D reconstruction method that incrementally updates a global point cloud from posed monocular RGB images. Our approach offers flexible local updates without the need for predefined resolutions or scene sizes, while ensuring surface consistency across views. A key contribution is the ray-based 2D-3D matching strategy, which improves feature matching robustness. Future work will focus on simplifying the architecture, reducing point cloud size, and refining the merging algorithm to enhance efficiency and performance.